\documentclass[11pt]{article}
\usepackage{coling2020}
\usepackage{times}
\usepackage{url}
\usepackage{latexsym}

\usepackage{graphicx}
\usepackage{bm} % zhu add
\usepackage{todonotes} % zhu add
\usepackage{makecell} % zhu add
\usepackage{amsmath,amsfonts,amssymb}%huang add 
\usepackage{multirow} % huang add fd
\newcommand{\mysizeTable}{\fontsize{7.8pt}{\baselineskip}\selectfont} %huang add
\title{Cycle-Consistent Adversarial Autoencoders \\ for Unsupervised Text Style Transfer}
\author{Yufang Huang$^1$ \space Wentao Zhu$^2$ \space Deyi Xiong$^3$ \space Yiye Zhang$^4$ \space Changjian Hu$^1$ \space Feiyu Xu$^1$ \\
  $^1$Lenovo Research AI Lab, $^2$Kwai Inc., $^3$Tianjin University, $^4$Cornell University  \\
  \{yfhuang1992new,wentaozhu91\}@gmail.com, dyxiong@tju.edu.cn \\ yiz2014@med.cornell.edu, \{hucj1,fxu\}@lenovo.com
}

\date{}

\begin{document}
\maketitle
\begin{abstract}
  Unsupervised text style transfer is full of challenges due to the lack of parallel data and difficulties in content preservation. In this paper, we propose a novel neural approach to unsupervised text style transfer which we refer to as Cycle-consistent  Adversarial autoEncoders (CAE) trained from non-parallel data. CAE consists of three essential components: (1)  LSTM autoencoders that encode a text in one style into its latent representation and decode an encoded representation into its original text or a transferred representation into a style-transferred text, (2) adversarial style transfer networks that use an adversarially trained generator to transform a latent representation in one style into a representation in another style, and (3) a cycle-consistent constraint that enhances the capacity of the adversarial style transfer networks in content preservation. The entire CAE with these three components can be trained end-to-end. Extensive experiments and in-depth analyses on two widely-used  public datasets consistently validate the effectiveness of proposed CAE in both style transfer and content preservation against several strong baselines in terms of four automatic evaluation metrics and human evaluation.
\end{abstract}

% \section{Credits}

% This document has been adapted from the instructions for  
% COLING-2018 proceedings compiled by Xiaodan Zhu and Zhiyuan Liu,
% which are, in turn, based on
% the instructions for
% COLING-2016 proceedings compiled by Hitoshi Isahara and Masao Utiyama,
% which are, in turn, based on
% the instructions for
% COLING-2014 proceedings compiled by Joachim Wagner, Liadh Kelly
% and Lorraine Goeuriot,
% which are, in turn, based on the instructions for earlier ACL proceedings,
% including 
% those for ACL-2014 by Alexander Koller and Yusuke Miyao,
% those for ACL-2012 by Maggie Li and Michael
% White, those for ACL-2010 by Jing-Shing Chang and Philipp Koehn,
% those for ACL-2008 by Johanna D. Moore, Simone Teufel, James Allan,
% and Sadaoki Furui, those for ACL-2005 by Hwee Tou Ng and Kemal
% Oflazer, those for ACL-2002 by Eugene Charniak and Dekang Lin, and
% earlier ACL and EACL formats. Those versions were written by several
% people, including John Chen, Henry S. Thompson and Donald
% Walker. Additional elements were taken from the formatting
% instructions of the {\em International Joint Conference on Artificial
%   Intelligence}.

\section{Introduction}
\label{intro}

%
% The following footnote without marker is needed for the camera-ready
% version of the paper.
% Comment out the instructions (first text) and uncomment the 8 lines
% under "final paper" for your variant of English.
% 
\blfootnote{
    %
    % for review submission
    %
    % \hspace{-0.65cm}  % space normally used by the marker
    % Place licence statement here for the camera-ready version. 
    %See Section~\ref{licence} of the instructions for preparing a manuscript.
    %
    % % final paper: en-uk version 
    %
    % \hspace{-0.65cm}  % space normally used by the marker
    % This work is licensed under a Creative Commons 
    % Attribution 4.0 International Licence.
    % Licence details:
    % \url{http://creativecommons.org/licenses/by/4.0/}.
    % 
    % % final paper: en-us version 
    %
    \hspace{-0.65cm}  % space normally used by the marker
    This work is licensed under a Creative Commons 
    Attribution 4.0 International License.
    License details:
    \url{http://creativecommons.org/licenses/by/4.0/}.
}

Unsupervised text style transfer is to rewrite a text in one style into a text in another style while the content of the text remains the same as much as possible without using any parallel data. Style transfer can be utilized in many tasks such as personalization in dialogue systems~\cite{oraby2018controlling,colombo-etal-2019-affect}, sentiment and word decipherment~\cite{shen2017style}, offensive language translation~\cite{santos2018fighting}, and data augmentation~\cite{perez2017effectiveness,8388338,zhu2020neurreg}, etc.

However, there are a variety of challenges to text style transfer in practice. First, we do not have large-scale style-to-style parallel data to train a text style transfer model in a supervised way. Second, even with non-parallel corpora, the inherent discrete structure of text sequences aggravates the difficulty of learning desirable continuous representations for style transfer~\cite{huang2020dice,pmlr-v80-zhao18b,bowman2016generating,hjelm2017boundary}. Third, it is difficult to preserve the content of a text when its style is transferred. To obtain good content preservation for text style transfer, various disentanglement approaches~\cite{shen2017style,hu2017toward,fu2018style,sudhakar-etal-2019-transforming} are proposed to separate the content and style of a text in the latent space. However, content-style disentanglement is not easily achievable as content and style typically interact with each other in texts in subtle ways~\cite{lample2018multiple}.

In order to solve the issues above, we propose a cycle-consistent adversarial autoencoders (CAE) for unsupervised text style transfer. In CAE, we learn the representation of a text where we embed both content and style in the same space. Such space is constructed for each style from non-parallel data. We then transfer the learned representation from one style space to another space. To guarantee that the content is preserved during the style transfer procedure, the transferred representation is transferred back to the original space to minimize the distance between its original representation and the reversely transferred representation.

\begin{figure}[t]
\centering
%\includegraphics[width=4.5cm]{model_AAAI20_190826V2.png}
%model_AAAI20_190829V1
\includegraphics[width=8cm]{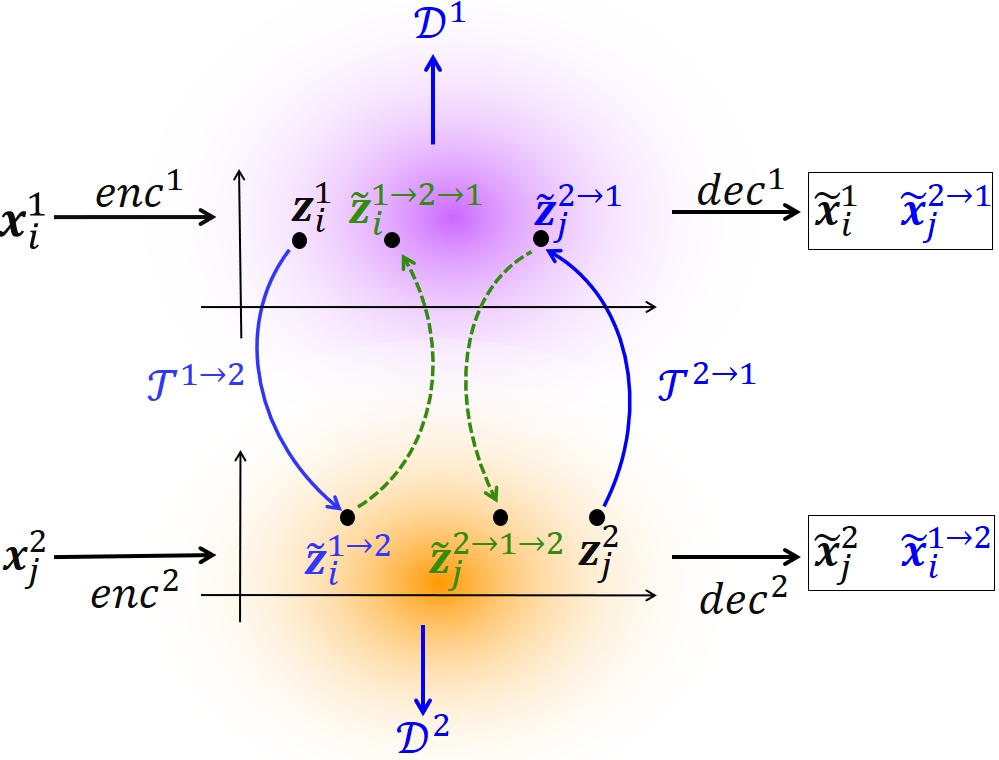}
\caption{The architecture of the proposed CAE network. 
%{\color{red}The CAE firstly employs LSTM autoencoders (${enc}^1, {dec}^1$ and ${enc}^2, {dec}^2$) to learn latent representations ($\bm{z}^1, \bm{z}^2$) for discrete text sequences. Then style transfers (${\mathcal{T}}^{1\rightarrow 2}$, ${\mathcal{T}}^{2\rightarrow 1}$) are conducted in the latent space and are trained based on cycle-consistent and generative adversarial losses. Finally, the CAE generates sentences of target styles from transferred representations by the decoders (${dec}^1, {dec}^2$) of LSTM autoencoders.}
} \label{fig:model}
\end{figure}

Figure~\ref{fig:model} demonstrates the diagram of the proposed CAE. Without loss of generality, we discuss CAE for text transfer between two styles. Multiple styles can be factorized into two styles~\cite{shen2020multi}. Multimodality Specifically, CAE is composed of three essential components: LSTM autoencoders, adversarial style transfer networks and a cycle-consistent constraint. The LSTM autoencoder contains an encoder $enc$ to encode a sentence $\bm{x}^s_i$ from style $s$ into a hidden representation $\bm{z}^s_i$ in the corresponding style space and a decoder $dec$ to generate sentences from vectors $\bm{z}^s_i$  learned by the LSTM encoder (or $\tilde{\bm{z}}_j$  transferred from the other style space). The adversarial style transfer networks learn a generator $\mathcal{T}$ to generate a representation $\tilde{\bm{z}}^{s_1\rightarrow s_2}_i$ in style space $s_2$ from $\bm{z}^{s_1}_i$ in style space $s_1$, or the other way around from style space $s_2$ to  $s_1$. It also uses a discriminator to ensure that the transferred representations belong to the corresponding style space.

\begin{figure}[t]
\centering
%\includegraphics[width=4.5cm]{model_AAAI20_190826V2.png}
%model_AAAI20_190829V1
\includegraphics[width=16cm]{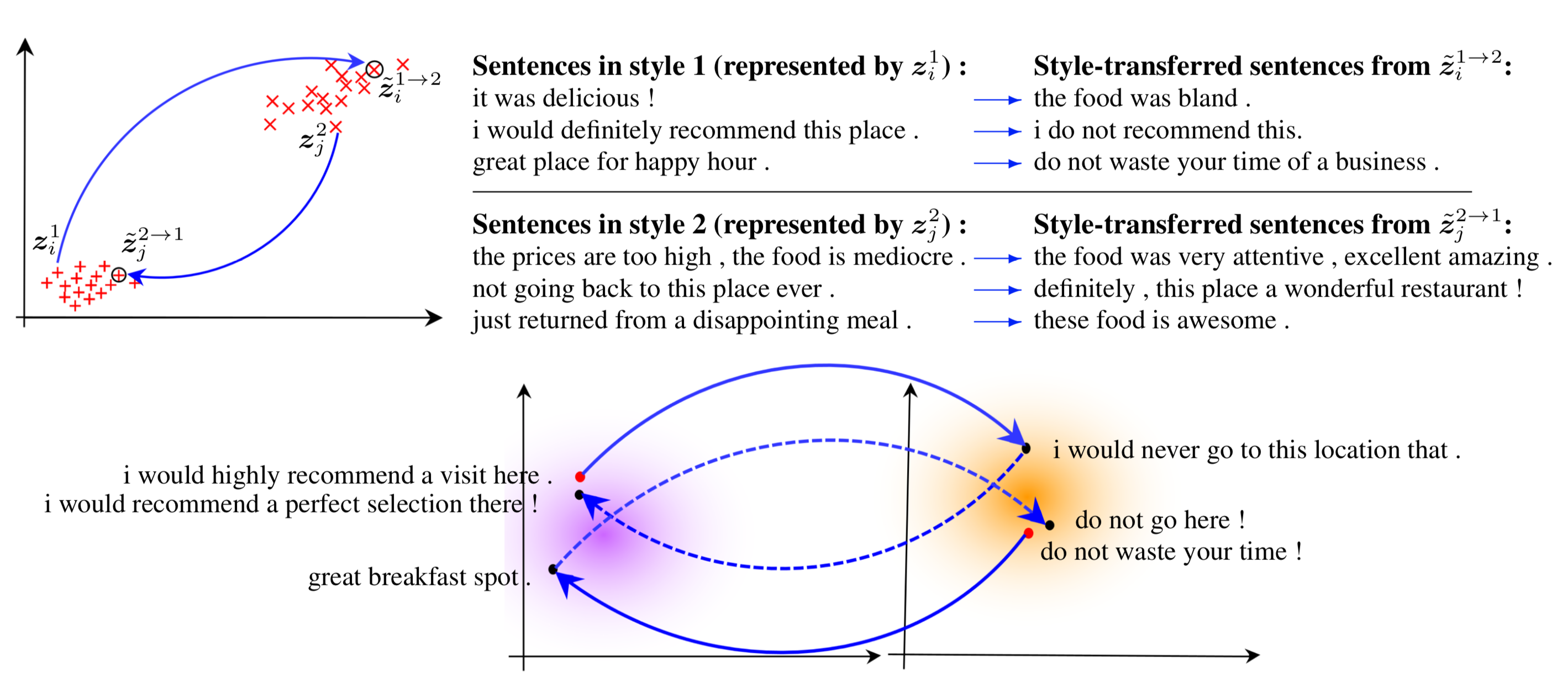}
\caption{The visualization of style transfer and cycle-consistent constraint in CAE. Upper: sentences in one style and style-transferred sentences in another style. Bottom: cycle-consistent constraint enforcing that sentences transferred into a different style can be translated back to its original meaning in its original style. 
} \label{fig:motivation}
\end{figure}

The top of Figure~\ref{fig:motivation} displays the original sentences and style-transferred sentences generated by the LSTM decoder from transferred representations produced by the generator of the adversarial style transfer network. The cycle-consistent constraint transfers back representations to their original space and attempts to minimize their distances, as demonstrated in the bottom of Figure~\ref{fig:motivation}. 

In summary, our contributions are threefold as follows.

\begin{itemize}
    \item We propose a novel end-to-end framework with three components to learn text style transfer without using parallel data.
    \item To the best of our knowledge, our work is the first to use the cycle-consistent constraint in the latent representational space for unsupervised text style transfer. 
    \item The proposed CAE are validated on two widely-used datasets: Yelp restaurant review sentiment transfer dataset and Yahoo QA topic transfer dataset. Extensive experiments and analyses demonstrate that CAE obtains better performance than several state-of-the-art baselines in both style transfer and content preservation. 
\end{itemize}

% \begin{itemize}%end-to-end trained
% \item We firstly employ a cycle-consistent constraint in the latent representational space for the generic style transfer task from non-parallel text. The latent representational cycle-consistency is validated to be effective as illustrated in Fig.~\ref{fig:motivation}\todo{ablation study for each of your claim}.
% %\vspace{-2mm}
% \item We design a novel deep generative network, Cycle-consistent Adversarial Regularized autoEncoder (CAE), for style transfer from non-parallel text as illustrated in Fig.~\ref{fig:model}. The CAE employs LSTM autoencoders to learn latent representations, conducts style transfers in the latent spaces, and generates transferred texts from the transferred latent representations using decoders from LSTM autoencoders.
% %\vspace{-2mm}
% \item The proposed CAE are validated on two public and widely used datasets, Yelp restaurant reviews sentiment transfer dataset and Yahoo questions topic transfer dataset. Extensive experiments demonstrate the effectiveness of proposed CAE. The CAE obtains better performance than previous approaches on the two datasets.
% \end{itemize}
%\vspace{-3mm}

\section{Related work}
%\section{Related Work}
% Style transfer from non-parallel data is an activate research topic in both computer vision and natural language processing. \citet{zhu2017unpaired} propose CycleGAN for unpaired image-to-image translation. The intrinsic difference between text and image is the discrete structure of text. To tackle representational challenge of discrete nature of text sequences, we enforce cycle-consistency on latent representation for text generation, where the latent representational spaces are learned by Seq2Seq models~\cite{sutskever2014sequence}.
A number of text style transfer approaches have been proposed in recent years following the pioneering study of style transfer in images \cite{gatys2015neural}. These approaches can be roughly categorized into two strands: methods that disentangle representations of style and content and the others that do not. 
%These approaches can be roughly categorized into two strands: methods that disentangle representations of style and content and the others that do not. 

% {\color{red}
In the first line of text style transfer, 
Hu et al.~\shortcite{hu2017toward} combine a variational autoencoder (VAE) with style discriminators to enforce that styles can be reliably inferred back from generated sentences.
%Cross-aligned auto-encoder
Shen et al.~\shortcite{shen2017style} uses discriminators to align hidden states of the transferred samples from one style with the true samples in the other to obtain the shared latent content distribution. 
Fu et al.~\shortcite{fu2018style} use an adversarial network to separate content representations from style representations.
%, then complete style transfer through style-dependent decoders or a style-embedding decoder model 
Prabhumoye et al.~\shortcite{prabhumoye2018style} fix the machine translation model and the encoder of the back-translation model to obtain content representations, then generate texts with classifier-guided style-specific generators.
Li et al.~\shortcite{li2018delete} extract content words by deleting style indicator words, then combine the content words with retrieved style words to construct the final output. 
%Cycled reinforcement learning~\cite{xu-etal-2018-unpaired}
Xu et al.~\shortcite{xu-etal-2018-unpaired} use reinforcement learning to jointly train a neutralization module which removes style words based on a classifier and an emotionalization module.
ARAE~\cite{pmlr-v80-zhao18b} and DAAE~\cite{shen2019educating} train GAN-regularized latent representations to obtain style-independent content representations, then decodes the content representations conditioned on style. He et al.~\shortcite{he2019probabilistic} presents a new probabilistic graphical model for unsupervised text style transfer. 
%an discrete autoencoder with  GAN-regularized latent representations to obtain style-independent content representations.             , and uses independency constraints reuse the VAE encoder as an additional discriminator for recognizing the content representations which can be recovered from the generated samples. 
% }                         , then complete style transfer through style-dependent decoders or a style-embedding decoder model.

% {\color{red}
In the second line of works that avoid disentangled representations of style and content,  Lample et al.~\shortcite{lample2018multiple} use back-translation technique on denoising autoencoder model with latent representation pooling to control the content preservation. Their experiments and analyses show that the content-style disentanglement is neither necessary nor always satisfied with practical requirements, even with the domain adversarial training that explicitly aims at learning disentangled representations. Style Transformer~\cite{dai-etal-2019-style} uses Transformer as a basic module to train a style transfer system. DualRL~\cite{Luo19DualRL} employs a dual reinforcement learning framework with two sequence-to-sequence models in two directions, using style classifier and back-transfer reconstruction probability as rewards. %result    
% }

We follow the second line and propose a novel method that makes no assumption on the latent representation disentanglement. But differently, we perform style transfer in the latent representational spaces of the source and target style. And inspired by CycleGAN~\cite{zhu2017unpaired,zhu2018adversarial} which uses a cycle loss on image style transfer to enforce the back-translation of a transferred image to be equivalent to the original image, we also impose a cycle-consistent constraint on our style transfer network. However, training style transfer networks with such a cycle constraint on discrete texts is quite different from those on images and non-trivial. In order to enable cycle training on texts, we project texts onto adversarially regularized latent space collectively learned by the LSTM autoencoders and adversarial transfer networks. Different from latent cross project with Euclidean distance for semi-supervised style transfer~\shortcite{shang2019semi}, we construct latent CycleGAN to generate high quality sentences for unsupervised style transfer.

\section{CAE: Cycle-consistent Adversarial Autoencoders}
%\vspace{-1mm}will give a detailed introduction to the
% In this section, we describe 
% proposed Cycle-consistent Adversarial Regularized autoEncoder (CAE).
Suppose we have two non-parallel text datasets $\mathbb{X}^1=\{\bm{x}^1_i\}_{i=1}^{n}$ and $\mathbb{X}^2=\{\bm{x}^2_j\}_{j=1}^{m}$ with different styles $s_1$ and $s_2$. The CAE employs LSTM autoencoder models to encode  discrete text sequences $\bm{x}^1_i, \bm{x}^2_j$ into representations $\bm{z}^1_i, \bm{z}^2_j$  , and to generate sentences $\tilde{\bm{x}}^{1 \rightarrow 2 }_i, \tilde{\bm{x}}^{2 \rightarrow 1 }_j$ based on latent representations $\tilde{\bm{z}}^{1\rightarrow 2  }_i, \tilde{\bm{z}}^{2\rightarrow 1}_j$ transferred by the adversarial transfer network $\mathcal{T}^{1\rightarrow 2}, \mathcal{T}^{2\rightarrow 1}$ from $\bm{z}^1_i, \bm{z}^2_j$.
%respectively.  
%The network architecture of the CAE consists of three parts: LSTM autoencoders, adversarial style transfer networks and cycle-consistent constraint.

% is illustrated in Fig.~\ref{fig:model}. It conducts latent representational adversarial generations and cycle-consistent reconstructions from latent transformations $\mathcal{T}^{1\rightarrow 2}, \mathcal{T}^{2\rightarrow 1}$.

\subsection{LSTM autoencoders} % Style
We use an LSTM \cite{hochreiter1997long} autoencoder to learn the latent representation of a text for each style. The encoder employs an LSTM recurrent neural network to map the input sequence to a latent representation with a fixed size, and the decoder utilizes the other LSTM network to generate an output sequence from a hidden representation~\cite{sutskever2014sequence}. 
%The output sequence can either be a reconstructed version of the input sequence or a sequence decoded from a style-transferred vector. For each input sequence, we add an end-of-sentence symbol ``$<$eos$>$" at the end of the sequence, and pad the sequence with a padding symbol ``$<$pad$>$" to ensure the same length $L$ for all input sequences. 
Given the $i$-$th$ input text sequence $\bm{x}^1_i=(x_{i,1}^1, x_{i,2}^1, \cdots, x_{i,L}^1)$ in style $s_1$, the LSTM autoencoder for style $s_1$  can be formulated as:
\begin{equation}\label{eq:lm}
p(\tilde{\bm{x}}_i^1  |{\bm{x}}^1_i;{enc}^1,{dec}^1) 
%=&p(\tilde{\bm{x}}_i^{1}|{enc}^1({\bm{x}}^1_i);{enc}^1, {dec}^1)\\
=\prod_{t=1}^{L}p(\tilde{x}^1_{i,t}|\bm{z}_i^1,\tilde{\bm{x}}_{i,<t}^1;{enc}^1, {dec}^1)
\end{equation}
where ${\tilde{\bm{x}}}^1_i=({\tilde{x}}_{i,1}^1,  \cdots, {\tilde{x}}_{i,L}^1)$ is the reconstructed sequence with the same length $L$ as
% the input sequence 
$\bm{x}^1_i$
, $\bm{z}^1_i={enc}^1(\bm{x}^1_i)$ is the learned latent representation from the encoder ${enc}^1$, $\tilde{\bm{x}}^1_{i,<t}$ are the tokens generated before $\tilde{x}^1_{i,t}$ and we start the decoder by a 
%special 
start-of-sentence symbol ``$<$bos$>$'' which is $\tilde{x}^1_{i,<1}$. $p(\tilde{x}_{i,t}^1|\bm{z}_i^1,\tilde{\bm{x}}^1_{i,<t};{enc}^1, {dec}^1)$ is the softmax output from decoder ${dec}^1$. For style $s_2$, similarly, we construct the other LSTM autoencoder with encoder ${enc}^2$ and decoder ${dec}^2$ to learn latent representation $\bm{z}^2_j$. 

The LSTM autoencoder tries to reconstruct the input sequence $\bm{x}^k_i$ with the output $\tilde{\bm{x}}^k_i$ from the networks ${enc}^k, {dec}^k$, where $k=1,2$ for different styles. The training objective function for the two LSTM autoencoders can be computed as:
\begin{equation}\label{eq:AEReconstructionLoss}
\!\mathcal{L}_{R}  ({enc}^1\!,\!{dec}^1\!,\!{enc}^2\!,\!{dec}^2) 
\!= \! - \!\frac{1}{n} \!\sum_{i=1}^{n} \!\log p(\tilde{\bm{x}}^1_i \!=\!\! {\bm{x}}^1_i|{\bm{x}}^1_i;\!{enc}^1\!,\!{dec}^1)\!- \!\frac{1}{m}\!\! \sum_{j=1}^{m}\! \log p(\tilde{\bm{x}}^2_j \!=\! {\bm{x}}^2_j|{\bm{x}}^2_j;\!{enc}^2\!,\!{dec}^2)\!\
\end{equation}
% \begin{equation}\label{eq:AEReconstructionLoss}
% \begin{aligned}
% &\mathcal{L}_{R}  ({enc}^1,{dec}^1,{enc}^2,{dec}^2) \\
% %=&  \mathcal{L}_{R}({enc}^1,{dec}^1) +\mathcal{L}_{R}({enc}^2,{dec}^2) \\
% = & - \frac{1}{n} \sum_{i=1}^{n} \log p(\tilde{\bm{x}}^1_i = {\bm{x}}^1_i|{\bm{x}}^1_i;{enc}^1,{dec}^1)- \frac{1}{m} \sum_{j=1}^{m} \log p(\tilde{\bm{x}}^2_j = {\bm{x}}^2_j|{\bm{x}}^2_j;{enc}^2,{dec}^2)
% \end{aligned}
% \end{equation}
The two LSTM autoencoders transform discrete sequences into latent continuous representations, which facilitate the style transfer models to perform style transfer and cycle training in the continuous space. % s
%\vspace{-2.5mm}

\subsection{Adversarial style transfer networks} 
%\vspace{-1.5mm}
% We construct transformation functions for latent representations in the style transfer. The latent representational transformation eliminates the challenge of discrete nature for text sequences~\cite{pmlr-v80-zhao18b}. And the latent representation generated from an LSTM autoencoder consists of semantically meaningful high level features which can be effective for style transfer. Our latent representational style transfer is constructed as:

Once we obtain the representations of text sequences in different styles via LSTM autoencoders, we learn two transformation functions $\mathcal{T}^{1\rightarrow 2}$ and $\mathcal{T}^{2\rightarrow 1}$ to map a representation in one style to the representation in the other style in the learned latent spaces. 
The style transfer in this way is formulated as: 
\begin{equation}\label{eq:transfer}
    \begin{aligned}
        \tilde{\bm{z}}^{1\rightarrow 2} = \mathcal{T}^{1\rightarrow 2}(\bm{z}^1), \quad
        \tilde{\bm{z}}^{2\rightarrow 1} = \mathcal{T}^{2\rightarrow 1}(\bm{z}^2)
    \end{aligned}
\end{equation}
where $\tilde{\bm{z}}^{1\rightarrow 2}$ is the generated latent representation in style $s_2$ from its original representation $z^1$ in style $s_1$ by the transformation $\mathcal{T}^{1\rightarrow 2}$, and $\tilde{\bm{z}}^{2\rightarrow 1 }$ is the generated latent representation  in style $s_1$ by 
%the transformation 
$\mathcal{T}^{2\rightarrow 1}$.

\par % leverage
We use generative adversarial networks~\cite{goodfellow2014generative} to learn the two transformation functions. Let's consider the learning of the transformation $\mathcal{T}^{1\rightarrow 2}$. We regard the function $\mathcal{T}^{1\rightarrow 2}$ as the generator that generates a representation in style $s_2$ from a representation in style $s_1$. We then build a discriminator $\mathcal{D}^2$ to distinguish representations in style $s_2$ from others. The generator tries to generate a representation that is able to fool the discriminator. The adversarial learning of the generator $\mathcal{T}^{1\rightarrow 2}$ and the discriminator $\mathcal{D}^2$ is formulated as: 
\begin{equation}\label{eq:AdversarialLoss1}
\begin{aligned} %\bm{z}_1,\bm{z}_2
 \!\min_{\mathcal{T}^{1\rightarrow 2}}  \max_{\mathcal{D}^{2}}  \mathcal{L}_{G}(\mathcal{T}^{1\rightarrow 2},\mathcal{D}^{2}) &=  \mathbb{E}_{\bm{z}^2 \sim {p_{\bm{z}^2}}} \left[\log \mathcal{D}^{2}(\bm{z}^2)\right] +\mathbb{E}_{\bm{z}^1 \sim {p_{\bm{z}^1}}} \left[\log(1-\mathcal{D}^2(\mathcal{T}^{1\rightarrow 2}(\bm{z}^1)))\right] \\
 =&  \mathbb{E}_{\bm{x}^2 \sim {p_{data}}}\! \!\left[\log \!\mathcal{D}^{2}\!({enc}^2 (\!\bm{x}^2))\right]\!\!+\!\mathbb{E}_{\bm{x}^1\! \sim {p_{data}}} \!\left[\log(\!1-\!\mathcal{D}^2\!(\mathcal{T}^{1\rightarrow 2}({enc}^1 (\!\bm{x}^1))))\right]
\end{aligned}
\end{equation}% (\bm{z}^2) (\bm{z}^1) (\bm{x}^2) (\bm{x}^1)

% \begin{equation}\label{eq:AdversarialLoss1}
% \begin{aligned} %\bm{z}_1,\bm{z}_2
%  &\min_{\mathcal{T}^{1\rightarrow 2}}  \max_{\mathcal{D}^{2}}  \mathcal{L}_{G}(\mathcal{T}^{1\rightarrow 2},\mathcal{D}^{2}) \\
%  =&  \mathbb{E}_{\bm{z}^2 \sim {p_{\bm{z}^2}}} \left[\log \mathcal{D}^{2}(\bm{z}^2)\right] +\mathbb{E}_{\bm{z}^1 \sim {p_{\bm{z}^1}}} \left[\log(1-\mathcal{D}^2(\mathcal{T}^{1\rightarrow 2}(\bm{z}^1)))\right] \\
%  =&  \mathbb{E}_{\bm{x}^2 \sim {p_{data}}} \left[\log \mathcal{D}^{2}({enc}^2 (\bm{x}^2))\right]+\mathbb{E}_{\bm{x}^1 \sim {p_{data}}} \left[\log(1-\mathcal{D}^2(\mathcal{T}^{1\rightarrow 2}({enc}^1 (\bm{x}^1))))\right]
% \end{aligned}
% \end{equation}% (\bm{z}^2) (\bm{z}^1) (\bm{x}^2) (\bm{x}^1)

 Similarly, we can derive the generative adversarial loss $ \mathcal{L}_{G}(\mathcal{T}^{2\rightarrow 1},\mathcal{D}^{1})$ for style transformation  function  $\mathcal{T}^{2\rightarrow 1}$ and discriminator $\mathcal{D}^1$. 

%  as follows:

% \begin{equation}\label{eq:AdversarialLoss2}
% \begin{aligned} %\bm{z}_1,\bm{z}_2
%  &\min_{\mathcal{T}^{2\rightarrow 1}}  \max_{\mathcal{D}^{1}}  \mathcal{L}_{G}(\mathcal{T}^{2\rightarrow 1},\mathcal{D}^{1}) \\ 
%  %=&  \mathbb{E}_{\bm{z}^1 \sim {p_{\bm{z}^1}}} \left[\log \mathcal{D}^{1}(\bm{z}^1)\right] +\mathbb{E}_{\bm{z}^2 \sim {p_{\bm{z}^2}}} \left[\log(1-\mathcal{D}^1(\mathcal{T}^{2\rightarrow 1}(\bm{z}^2)))\right] \\
% =
% &  \mathbb{E}_{\bm{x}^1 \sim {p_{data}}} \left[\log \mathcal{D}^{1}({enc}^1 (\bm{x}^1))\right] +\mathbb{E}_{\bm{x}^2 \sim {p_{data}}} \left[\log(1-\mathcal{D}^1(\mathcal{T}^{2\rightarrow 1}({enc}^2 (\bm{x}^2))))\right]
% \end{aligned}
% \end{equation}%(\bm{z}^1) (\bm{z}^2) (\bm{x}^1) (\bm{x}^2)

\subsection{Cycle-consistent constraint}
Theoretically, the adversarial style transfer networks described above are capable of learning many different transformation functions that can generate outputs in the distribution identical to the target style space~\cite{zhu2017unpaired}. This is because that the learning of the transformation functions lacks of sufficient constraints and the two functions are learned in a relatively separate way according to equations (4) and (5).

% Theoretically, the generative adversarial training can learn mappings $\mathcal{T}^{2\rightarrow 1}$ and $\mathcal{T}^{1\rightarrow 2}$ that produce outputs identically distributed as target latent domains $\bm{z}^1$ and $\bm{z}^2$, respectively. However, a transfer network can map the same set of source latent representations to any random permutation of latent representations in the target latent domain. From the generative adversarial training, any of the learned mappings can induce an output distribution that matches the target latent distribution because of large capacity of the transfer network. Thus, generative adversarial loss alone cannot guarantee that the learned function can map an individual latent input $\bm{z}^1$ to a desired latent output $\tilde{\bm{z}}^{1\rightarrow 2}$ typically~\cite{zhu2017unpaired}. % a transfer network can map the same set of source latent representations to any random permutation of latent representations in the target latent domain where any of the learned mappings can induce an output distribution that matches the target latent distribution because of large capacity of the transfer network. ~\cite{goodfellow2014generative}

In order to learn desirable transformation functions, we use a cycle-consistent constraint to tighten the learning of the two transformation functions $\mathcal{T}^{1\rightarrow 2}$ and $\mathcal{T}^{2\rightarrow 1}$, which is inspired by CycleGAN~\cite{zhu2017unpaired}. 
The cycle-consistent constraint expects that a transferred representation generated by a transformation function can be translated back to its original representation by the other transformation function. 

Given a latent representation $\bm{z}^1$ in style $s_1$, the reconstructed latent representation through the two style transformation functions $\mathcal{T}^{1\rightarrow 2}, \mathcal{T}^{2\rightarrow 1}$ can be obtained as:
\begin{equation}
\tilde{\bm{z}}^{(1\rightarrow 2) \rightarrow 1 } =  \mathcal{T}^{2\rightarrow 1}(\tilde{\bm{z}}^{1\rightarrow 2}) = \mathcal{T}^{2\rightarrow 1}( \mathcal{T}^{1\rightarrow 2} (\bm{z}^{1}))
\end{equation}
%where $\tilde{\bm{z}}^{(1\rightarrow 2) \rightarrow 1 }$ is the reconstructed latent representation. 
Similarly, we can obtain the reconstructed latent representation $\tilde{\bm{z}}^{(2\rightarrow 1) \rightarrow 2 }$ for latent representation $\bm{z}^2$ in style $s_2$.
% :
% \begin{equation}
% \tilde{\bm{z}}^{(2\rightarrow 1) \rightarrow 2 } =  \mathcal{T}^{1\rightarrow 2}(\tilde{\bm{z}}^{2\rightarrow 1}) = \mathcal{T}^{1\rightarrow 2}( \mathcal{T}^{2\rightarrow 1} (\bm{z}^{2}))
% \end{equation} % 

To constrain the transformation functions $\mathcal{T}^{2\rightarrow 1}$ and $\mathcal{T}^{1\rightarrow 2}$, the latent representational cycle-consistent reconstruction loss can be formulated as:
\begin{equation}\label{eq:CycleConsistentReconLoss}
\begin{aligned}
\mathcal{L}_{C} (\mathcal{T}^{2\rightarrow 1},\mathcal{T}^{1\rightarrow 2})= &\mathbb{E}_{\bm{z}^1\sim p_{\bm{z}^1}}\left[\|\tilde{\bm{z}}^{( 1\rightarrow 2)\rightarrow 1 }-\bm{z}^1\|_1\right]+\mathbb{E}_{\bm{z}^2\sim p_{\bm{z}^2}}\left[\|\tilde{\bm{z}}^{(2\rightarrow 1)\rightarrow 2 }-\bm{z}^2\|_1\right] \\
=& \mathbb{E}_{\bm{x}^1\sim p_{data}}\left[\|\mathcal{T}^{2\rightarrow 1}( \mathcal{T}^{1\rightarrow 2} ( {enc}^1 (\bm{x}^1) ) )-{enc}^1 (\bm{x}^1)\|_1\right]
\\
&+\mathbb{E}_{\bm{x}^2\sim p_{data}}\left[\| \mathcal{T}^{1\rightarrow 2} ( \mathcal{T}^{2\rightarrow 1} ( {enc}^2 (\bm{x}^2) ) ) - {enc}^2 (\bm{x}^2)\|_1\right]
\end{aligned}%\tilde{\bm{z}}_{1\leftarrow (1\rightarrow 2 )}  \tilde{\bm{z}}_{2\leftarrow (2\rightarrow 1)}
\end{equation} % (\bm{z}^1) (\bm{z}^2) (\bm{x}^1) (\bm{x}^2)
where $\|\cdot\|_1$ is $L_1$ norm.

This latent representational cycle-consistent reconstruction imposes the constraint on the adversarial style transfer networks to palliate mode-dropping in the latent style transfer, and to improve the content preservation in the generated sentences. % as illustrated in Fig.~\ref{fig:motivation}. %We know that the content preservation is one of the most difficult task for style transfer of non-parallel text. 

\subsection{Training and inference}
As CAE has three components in its network architecture, the end-to-end training objective of CAE is composed of three sub-objectives and is formulated as: 
% The proposed Cycle-consistent Adversarial Regularized autoencoder (CAE) consists of three parts, LSTM autoencoders, latent representational generative adversarial networks (GANs) and latent representational cycle-consistency. The training objective function for the CAE can be written as: %\todo{r u use lambda1 lambda2 in your implement?}\todo{\color{purple}Yes but No, there no lambda change in my implement, $\lambda_1$ is fixed to be 0.1, and i just change the learning rate for different experiment.}
\begin{equation}\label{eq:CAELoss}
%&\mathcal{L}_{CAE}({enc}^1,{dec}^1,{enc}^2,{dec}^2,\mathcal{T}^{2\rightarrow 1},\mathcal{T}^{1\rightarrow 2},\mathcal{D}^1,\mathcal{D}^2)
%=&\mathcal{L}(D_1,D_2,E_1,F_1,E_2,F_2,T_{12},T_{21})\\
%\\=
\mathcal{L}_{CAE}\!=\!\lambda_1\!\mathcal{L}_{R}({enc}^1\!,\!{dec}^1\!,\!{enc}^2\!,\!{dec}^2)\!+\! \lambda_2\!\left(\mathcal{L}_{G}(\mathcal{T}^{2\rightarrow 1},\mathcal{D}^{1})\!+\!\mathcal{L}_{G}(\mathcal{T}^{1\rightarrow 2},\mathcal{D}^{2})\right)
\!+\!\lambda_3\!\mathcal{L}_{C}(\mathcal{T}^{2\rightarrow 1},\mathcal{T}^{1\rightarrow 2})
\end{equation}
where $\lambda_1$, $\lambda_2$ and $\lambda_3$ control the relative importance of the three sub-objectives. We aim to solve: 
\begin{equation}
{enc}^1,{dec}^1,{enc}^2,{dec}^2,\mathcal{T}^{2\rightarrow 1},\mathcal{T}^{1\rightarrow 2},\mathcal{D}^1,\mathcal{D}^2= \arg\min_{\{{enc}^1,{dec}^1,{enc}^2,{dec}^2,\mathcal{T}^{2\rightarrow 1},\mathcal{T}^{1\rightarrow 2}\}}\max_{\{\mathcal{D}^1,\mathcal{D}^2\}}\mathcal{L}_{CAE}
\end{equation}

For the inference, let's consider the transfer of a text $\bm{x}^1_i$ in style $s_1$ into a text in style $s_2$. We first obtain latent representation $\bm{z}_i^1={enc}^1 (\bm{x}_i^1)$ using encoder ${enc}^1$. We then perform style transfer and obtain the transferred latent representation $\tilde{\bm{z}}^{1\rightarrow 2 }_i$ in style $s_2$ based on equation (\ref{eq:transfer}). Finally, we employ the decoder ${dec}^2$ to generate a transferred sequence $\tilde{\bm{x}}^{1\rightarrow 2 }_i \sim {dec}^2 (\tilde{\bm{z}}_i^{1\rightarrow 2 })$ in style $s_2$:%\todo{should be arg max?}
\begin{equation}
{\tilde{x}^{1\rightarrow 2}_{i,t}} = \mathop{\arg\max}_{\tilde{x}^{1\rightarrow 2}_{i,t}} p(\tilde{x}_{i,t}^{1\rightarrow 2 }  | \tilde{\bm{z}}^{1\rightarrow 2 }, \tilde{\bm{x}}^{1\rightarrow 2  }_{i,<t}; {dec}^2)
\end{equation}
where $\tilde{\bm{x}}^{1\rightarrow 2 }_i = ({\tilde{x}^{1\rightarrow 2}}_{i,1}, \cdots, {\tilde{x}^{1\rightarrow 2 }_{i, L'}})$ with length $L'$,  $p(\cdot | \bm{z}, \cdot; {dec}^2)$ is the same as equation (1) calculated
by the softmax from $ {dec}^2$ with previous tokens. %The final generated sentence is truncated by the first end-of-sentence token ``$<$eos$>$''. 
The inference of the entire sequence $\tilde{\bm{x}}_i^{1\rightarrow 2 }$ in style $s_2$ from sequence $\bm{x}^1_i$ in style $s_1$  is formulated as:
\begin{equation}
\begin{aligned}
    \tilde{\bm{x}}^{1\rightarrow 2 }_i &\sim {dec}^2(\tilde{\bm{z}}_i^{1\rightarrow 2 }) = {dec}^2(\mathcal{T}^{1\rightarrow 2} (\bm{z}_i^1)) 
    = {dec}^2(\mathcal{T}^{1\rightarrow 2} ( {enc}^1 (\bm{x}_i^1)))
\end{aligned}
\end{equation}
Similarly, we can conduct style transfer from a sequence $\bm{x}^2_j$ in style $s_2$ to generate a sequence $\tilde{\bm{x}}^{2\rightarrow 1}_j$ in style $s_1$.
% :
% \begin{equation}
% \begin{aligned}
%     \tilde{\bm{x}}^{2\rightarrow 1}_j &\sim {dec}^1(\tilde{\bm{z}}_j^{2\rightarrow 1}) = {dec}^1(\mathcal{T}^{2\rightarrow 1} (\bm{z}_j^2)) 
%     = {dec}^1(\mathcal{T}^{2\rightarrow 1} ( {enc}^2 (\bm{x}_j^2)))
% \end{aligned}
% \end{equation}
%In summary, our full objective is 

\section{Experiments}
%\section{EXPERIMENTS}
%\subsection{Datasets}
To compare our work with previous approaches to text  style transfer from non-parallel data, we conducted experiments on two text transfer tasks: sentiment transfer on the Yelp restaurant review corpus     and topic transfer on the ``Yahoo! Answers Comprehensive Questions and Answers version 1.0" dataset. We also carried out ablation experiments to study the impact of different components of CAE on overall performance of style transfer. %  
%\footnote{www.yelp.com/dataset}  
%\footnote{webscope.sandbox.yahoo.com/catalog.php?datatype=l}

\subsection{Experimental setup}
\subsubsection{Datasets} For the Yelp dataset, we followed the same experimental setup and used the same dataset as Cross-aligned auto-encoder~\cite{shen2017style} and ARAE~\cite{pmlr-v80-zhao18b} for sentiment transfer on the Yelp restaurant reviews. The sentiment of a review is labeled as positive if the rating is above three; otherwise, it is labeled as negative. We used  $70\%$ of the data for training, $10\%$ for validation and the rest for testing.

For the Yahoo QA dataset, we chose two topics for style transfer: ``Entertainment $\&$ Music" and ``Politics $\&$ Government", and extracted questions from these two topics to construct the final dataset. The  partition ratios of this dataset for training and testing are $80\%$ and $20\%$, respectively.  
To reduce the vocabulary size, we pruned the vocabulary to keep the most frequent words and replaced other words with ``$<$unk$>$". Table~\ref{table_vocabulary} shows the statistics of the two datasets. 

\begin{table}[t]
\small
\begin{center}
\begin{tabular}{c|c|c|c|c}
\Xhline{2\arrayrulewidth} \bf Dataset &\multicolumn{2}{c|}{Yelp}&\multicolumn{2}{c}{Yahoo}\\ \hline
%Styles &Pos.&Neg.& \begin{tabular}{@{}c@{}}Entertain. \\ $\&$Music\end{tabular} & \begin{tabular}{@{}c@{}}Politics  \\ $\&$Govern.\end{tabular} \\  

Styles & Positive &Negative &Entertainment $\&$ Music & Politics $\&$ Government \\ \Xhline{2\arrayrulewidth}
%&&&&$\&$Politics \\ \hline 
$\#$Sent. & 382K & 252K & 441K & 153K\\ \hline
$\#$Vocab.&\multicolumn{2}{c|}{10K}&\multicolumn{2}{c}{116K} \\ \hline
%\begin{tabular}{@{}c@{}}$\#$Pruned \\ vocab.\end{tabular} &\multicolumn{2}{c|}{10K}&\multicolumn{2}{c}{30K} \\ \Xhline{2\arrayrulewidth}
$\#$Pruned Vocab.&\multicolumn{2}{c|}{10K}&\multicolumn{2}{c}{30K} \\ \hline

\end{tabular}
\end{center}
\caption{\label{table_vocabulary}Statistics of Yelp review sentiment transfer dataset and Yahoo {QA} topic transfer dataset.}
\end{table}
%\subsection{Results and Discussion}

\subsubsection{Baselines}
We compared CAE with the following baselines: 
% \begin{itemize}
% \item 
(1) \textbf{LSTM autoencoder (AE)}: using only LSTM autoencoders in CAE for the two styles without the style transfer networks and cycle-consistent constraint. 
% \item 
(2) \textbf{Cross-aligned autoencoder (Cross-aligned AE)}~\cite{shen2017style}: aligning the hidden states of autoencoders adversarially to learn a shared latent content distribution.
% \item 
(3) \textbf{ARAE}~\cite{pmlr-v80-zhao18b}: adversarially training GAN-regularized prior with a classifier to obtain style-independent content representations, then conducting style transfer through decoders conditioned on style.
% \item 
(4) \textbf{Template-based method}~\cite{li2018delete}: replacing the style words of source sentence with the other style words retrieved from target sentences. 
% \item 
(5) \textbf{Cycled reinforcement learning approach (Cycled RL)}~\cite{xu-etal-2018-unpaired}: using reinforcement learning to jointly train a neutralization module which removes style words based on a classifier and an emotionalization module.% which add style information by multi-decoder.%~\cite{xu-etal-2018-unpaired}. 
% \end{itemize}
%The results of AE, Cross-aligned AE, ARAE, and Cycled RL on Yelp dataset are reported according to their papers. For the template-based method and experiments on Yahoo dataset, we rerun their systems on our datasets to obtain the results. 
%\vspace{-2.5mm}
\subsubsection{Hyper-parameter settings} %sentiment transfer topic transfer
The used encoders ${enc}^1, {enc}^2$ and decoders ${dec}^1, {dec}^2$ were LSTM networks with one hidden layer of size $h_n=128$ on the Yelp review dataset and of size $h_n = 300$ on Yahoo QA dataset. The word embedding size was the same as the number of hidden neurons $h_n$. The latent variables $\bm{z}^1, \bm{z}^2$, $\tilde{\bm{z}}^{2\rightarrow 1}$ and $\tilde{\bm{z}}^{1\rightarrow 2 }$ were $L_2$-normalized to $1$. 
%The length $L$ was fixed for sequences in the pre-processing and set as the maximal length of the sequences in each batch. The maximal length $L'$ of sequences generated by decoders ${dec}^1, {dec}^2$ was set as $50$ in all the experiments. 
The transformation functions $\mathcal{T}^{1\rightarrow 2}, \mathcal{T}^{2\rightarrow 1}$ were parameterized by two-layer fully-connected neural networks ($h_n$-$h_n$-$h_n$ neurons). The discriminators $\mathcal{D}^1, \mathcal{D}^2$ were two-layer fully-connected neural networks ($h_n$-$h_n$-$1$ neurons) with hyperbolic tangent activation function in the first layer and sigmoid activation function for the second layer. The $\lambda_1,\lambda_2,\lambda_3$ were set as $0.1$, $1.0$, $1.0$ respectively based on the performance on the validation set.
%\vspace{-2.5mm} 

\subsubsection{Evaluation metrics}
%\vspace{-2mm}

We used four automatic metrics to quantitatively evaluate the proposed CAE: {Transfer}, {BLEU}, {PPL} and {RPPL}, which have been widely used in previous literature~\cite{pmlr-v80-zhao18b}. {Transfer} is the style transfer success rate and implemented as a classifier which is trained by the {\emph{fastText}} library~\cite{joulin2016bag}. {BLEU} is used to evaluate the content preservation between the source sequence and transferred sequence~\cite{papineni2002bleu}. To evaluate th{}e fluency of the transferred sequence, we utilized the perplexity of the generated text denoted by {PPL}. We also used the reverse perplexity ({RPPL}) to assess the representativeness of generated texts with respect to the underlying data distribution and to detect the mode collapse for generative models~\cite{pmlr-v80-zhao18b}. RPPL scores were calculated by training an RNN language model on generated samples to evaluate the perplexity on real-world hold-out data~\cite{pmlr-v80-zhao18b}. We used the code from Zhao et al.~\cite{pmlr-v80-zhao18b} (word embedding size of $300$ with dropout $0.2$, and one-layer LSTM of size $300$ with dropout $0.2$) to build the language models and calculate {PPL} and {RPPL}. These four evaluation metrics together form a comprehensive evaluation and comparison between different approaches. 

We also conducted human evaluation. We randomly chose 200 instances from each style for the human evaluation. Four human annotators can proficiently understand English texts and have sufficient background knowledge about this evaluation task. The annotation is blind to them in random order. They grade all sentences with scores from one to five for style transfer, content preservation and fluency. Following 
%Point-Then-Operate (PTO)~\cite{WuRLS19} and template-based method~\cite{li2018delete}
Wu et al.~\shortcite{WuRLS19} and Li et al.~\shortcite{li2018delete}, we regard a style transfer with scores larger than or equal to four on all three measures (style transfer, content preservation and fluency) as a successful transfer. We calculate the percentage of successful transfers and refer to this percentage as Suc.

\begin{table}
\begin{center}
\begin{tabular}{c|cccc|cccc}
\Xhline{2\arrayrulewidth} \bf Dataset &\multicolumn{4}{c|}{Yelp}&\multicolumn{4}{c}{Yahoo}\\ 
\Xhline{2\arrayrulewidth} \bf Models & \bf Transfer$\uparrow$ & \bf BLEU$\uparrow$ &\bf PPL $\downarrow$&\bf RPPL$\downarrow$ & \bf Transfer & \bf BLEU  &\bf  PPL  &\bf RPPL \\ \Xhline{2\arrayrulewidth}
AE &$59.3\%$& 37.28&31.9&68.9  &$62.0\%$& 32.2&62.7&57.3 \\ \hline 
%{\color{red}TemplateBased({\color{red}{business}}) }& $81.7\%$ & 11.8 & & \\ \hline 
{Template} & $80.2\%$ & \bf 52.90 &161.1&64.0& $\textbf{85.3\%}$ & \bf 58.9 &137.5&178.4\\ \hline 
{Cycled RL} &  $81.4\%$ & 22.00 &81.9 & 85.6  &  $75.0\%$ & 20.2 & {37.5}& {305.7}\\ \hline 
Cross-aligned AE &$77.1\%$ &17.75&65.9&124.2 &$73.4\%$ &11.5&36.5& 79.3\\ \hline

 %ARAE$_{1}$   & $73.4\%$& 31.15& 29.7& 70.1\\ \hline
 ARAE   &$81.8\%$ & 20.18&27.7 &77.0 & $77.7\%$& 21.1& 34.4& 53.7\\ \Xhline{2\arrayrulewidth}
 %ARAE&$81.5\%$&18.72&28.1&\textbf{52.2} \\\Xhline{2\arrayrulewidth}
 %(yfv27) 
 %CAE$_1$ &$73.7\%$ & {33.41}  & 25.5& \textbf{47.1}\\ \hline
%(yfv14) 
CAE&$\textbf{86.9\%}$  &22.51 &\textbf{21.6} & \bf 57.0&$ 83.8\% $  &   { 23.0} & \bf 25.4 & \bf 52.6\\ 
\Xhline{2\arrayrulewidth}
\end{tabular}
\end{center}
\caption{\label{table_1} Quantitative comparisons. Left: results on Yelp restaurant review sentiment transfer dataset, right: result on Yahoo QA topic transfer dataset.}
\end{table}

\subsection{Results}
\subsubsection{Yelp restaurant reviews sentiment transfer}
% We firstly conduct sentiment transfer on Yelp restaurant reviews dataset and compare CAE with AE, template based method~\cite{li2018delete}, Cycled RL~\cite{xu-etal-2018-unpaired}, cross-aligned AE~\cite{shen2017style} and ARAE~\cite{pmlr-v80-zhao18b}. From Table~\ref{table_1}, 
The results are shown in Table~\ref{table_1} (left), from which we clearly observe that CAE obtains better performance than the five baseline approaches. Specifically, CAE yields improvements of $5.1$, $6.1$ and $7.0$ points over the best baselines for sentiment transfer in terms of transfer success rate ({Transfer}), fluency ({PPL}) and mode reservation ({RPPL}), respectively.

%with good content preservation ({BLEU}) simultaneously. 
% Because CAE employs different LSTM autoencoders for text sequences of different styles, it obtains better {PPL}, which indicates that the transferred sequence is more fluent than the compared approaches. The latent representational cycle-consistency on transfer functions palliates mode-dropping in the style transfer and facilitates to produce low {RPPL}. The latent generative adversarial and cycle-consistent losses improve the content preservation ({BLEU}) and transfer ability simultaneously. The transfer success rate of AE is undesirable and the template based method has low fluency ({PPL}), although the two methods reserves more content ({BLEU}).
The template-based method achieves the highest BLEU score since all content words are guaranteed to be kept by templates with only style words replaced by retrieved words. However, it obtains the worst perplexity, indicating that it is very difficult for the template-based method to generate fluent sentences. By contrast, our CAE achieves the lowest perplexity due to the strong LSTM decoders. The cycle-consistent constraint enables CAE to yield the best RPPL as it palliates mode dropping in style transfer. Additionally, the adversarial style transfer network constrained by the cycle consistency loss facilitates CAE to perform well on both style transfer and content preservation. 

%The ARAE$_1$  and ARAE$_{2}$ are ARAEs with two different weights (1 and 10) on the generative adversarial losses, respectively, and their results are reported from~\citet{pmlr-v80-zhao18b}. The ARAE$_{2}$ enhances transfer by a large generative adversarial loss, but tends to generate sequence less similar to the source than ARAE$_{1}$ which can be found from the low value of {\bf BLEU} score in Table~\ref{table_1}. We also conduct two experiments based on different training procedures: 1) CAE$_1$: we firstly train autoencoder Seq2Seq networks, then use equation (\ref{eq:CAELoss}) to train the CAE. 2) CAE$_2$: we firstly train autoencoder Seq2Seq networks, then employ generative adversarial loss to train transfer networks, and finally use equation (\ref{eq:CAELoss}) to train the CAE. We conclude the quantitative comparison results in Table~\ref{table_1}.

%\vspace{-2.2mm}
\subsubsection{Yahoo questions topic transfer}
%\vspace{-1.8mm}
We further evaluated CAE against five basedlines on the Yahoo QA topic transfer task. Results in Table~\ref{table_1} (right) show that CAE obtains better {Transfer} ($+6.1\%$), {PPL} ($-9$), {BLEU} and {RPPL} than the best baseline approaches of Cycled RL, Cross-aligned AE and ARAE, demonstrating the advantages of CAE on style transfer, fluency and content preservation. 

The template-based method again achieves the highest BLEU score but it fails to generate meaningful and fluent sentences (very high PPL and RPPL). The reason for this is because simply replacing the style words with unreasonable words may generate senmantically incorrect sentences. For example, the template-based method transfers a source sentence of ``is harrison ford married ?" into ``is a state in the married ?", which is meaningless. The template-based method also achieves the highest transfer which is different with the results on the Yelp dataset. The reason is that it is easier to differentiate style words from content words in the Yahoo topic dataset than the Yelp dataset which makes it more accurate for the template-based method to substitute style words. It can be observed that the template-based method achieves high BLEU scores at the cost of fluency and semantic correctness. Taking all the four metrics into consideration, we believe that our approach performs better than all the baselines.

% The results validate CAE yields good style transfer, fluency, content preservation simultaneously. It demonstrates the effectiveness of latent representational cycle-consistent generative adversarial training for semantic style transfer.

%Here we don't use Cycled RL method as baseline, since this method is used for sentiment-to-sentiment translation, not available here for topic transfer. We also don't use baseline AE and Template because they get bad Transfer and Fluency respectively according to last experiment. 
% The quantitative comparison results are concluded in Table~\ref{table_6}. 

% From Table~\ref{table_6}, we can find that the CAE obtains better {Transfer} ($2.5\%$), {PPL} ($9.5$) and {BLEU} than the ARAE. The results validate the CAE yields good style transfer, fluency, content preservation simultaneously. It demonstrates the effectiveness of latent representational cycle-consistent generative adversarial training for semantic style transfer. %\todo{result of CAE$_1$? You need to do in case reviews}. %, at the same time we ensure the content preserving.

\subsubsection{Human evaluation} 

Table~\ref{table_humanEvaluation} shows the results of human evaluation. The CAE achieves the highest style transfer, content preservation and fluency score on both datasets. It also obtains the highest comprehensive successful rate of style transfer in terms of Suc. 

{
\begin{table}
\begin{center}

\begin{tabular}{c||ccc|c||ccc|c}
\Xhline{2\arrayrulewidth} \bf Dataset &\multicolumn{4}{c||}{Yelp}&\multicolumn{4}{c}{Yahoo}\\ 
 \Xhline{2\arrayrulewidth}\bf Models & \bf Style $\uparrow$ & \bf Content $\uparrow$  &\bf Fluency $\uparrow$  &\bf Suc $\uparrow$& \bf Style & \bf Content  &\bf Fluency  &\bf Suc  \\ \Xhline{2\arrayrulewidth}
Cross-aligned AE& 4.09 & 3.99 &4.14&52$\%$ & 3.90 & 3.90  &3.85&40$\%$\\  
ARAE &4.03  & 4.15 &4.36& 55$\%$ & 3.99  & 4.10  & 4.25 & 51$\%$\\ 
CAE & \bf 4.19  & \bf 4.23  & \bf 4.45&\bf 65$\%$ & \bf 4.16 & \bf 4.20 & \bf 4.34&\bf 63$\%$ \\
\Xhline{2\arrayrulewidth}
\end{tabular}

% \begin{tabular}{c|ccc|c}
% \Xhline{2\arrayrulewidth} \bf Models & \bf Style & \bf Content  &\bf Fluency  &\bf Suc \\ \Xhline{2\arrayrulewidth}
% Cross-aligned AE& 4.09 & 3.99 &4.14&52$\%$\\  \hline
% ARAE &4.03  & 4.15 &4.36& 55$\%$\\ \Xhline{2\arrayrulewidth}
% CAE &4.19  &4.23  &4.45&\bf 65$\%$\\
% \Xhline{2\arrayrulewidth}
% \end{tabular}

% \begin{tabular}{c|ccc|c}
% \Xhline{2\arrayrulewidth} \bf Models & \bf Style & \bf Content  &\bf Fluency  &\bf Suc \\ \Xhline{2\arrayrulewidth}
% Cross-aligned AE& 3.90 & 3.90  &3.85&40$\%$\\ \hline
% ARAE & 3.99  & 4.10  & 4.25 & 51$\%$\\ \Xhline{2\arrayrulewidth}
% CAE & 4.16 & 4.20 &4.34&\bf 63$\%$\\
% \Xhline{2\arrayrulewidth}
% \end{tabular}
\end{center}
\caption{\label{table_humanEvaluation}Human evaluation. Left: Yelp dataset. Right: Yahoo dataset.}
\end{table}}

{ \begin{table}

\begin{center}
\begin{tabular}{ccccc}
\Xhline{2\arrayrulewidth} \bf Models & \bf Transfer $\uparrow$ & \bf BLEU $\uparrow$  &\bf PPL $\downarrow$  &\bf RPPL $\downarrow$ \\ \Xhline{2\arrayrulewidth}

CAE & $86.9\%$ & 22.51 &21.6&57.0\\  \Xhline{2\arrayrulewidth}

w/o cycle-consistency& $88.2\%$ & 14.5 & 24.4 & 60.6 \\ 
w/o discriminators& $99.2\%$ & 0.2 & 25.7 & 701.1\\ 
\Xhline{2\arrayrulewidth}
\end{tabular}
\end{center}
\caption{\label{table_ablation}Ablation study on Yelp dataset.}
\end{table}}

\subsection{Ablation study} % restaurant reviews sentiment transfer on overall performance . We conduct two ablations, one is the the other is it's also possible to 
We further conducted ablation experiments on the Yelp dataset to study the effect of the cycle-consistent constraint and discriminators in CAE. Table~\ref{table_ablation} shows the results. When we disable the cycle-consistent constraint, we can train the model successfully. However, it leads to significant drop in BLEU and higher PPL and RPPL with marginal improvement in Transfer compared to the full CAE, which again confirms that the cycle-consistent constraint is helpful for content {mode} preservation. 
When we disable the discriminators, the model cannot preserve content, obtaining terribly low BLEU and high RPPL, indicating complete mode collapse. Without the discriminators, the CAE has no constraint and guidance to learn the transformation functions, which is prone to mode collapse. Serious mode collapse results in poor content preservation. We notice that most transferred sentences in the ``negative" style contain words ``not" or ``disappointed", while most generated sentences in the ``positive" style have the word ``good". The discriminators are important for preventing the model from collapse and hence further preserving the content.
%The generated sentences lack diverse style words. 
%CAE without discriminators fails because the styles words in generated sentences lack diversity. %{\color{red}The generated sentences lack diversity} and the style transfer without the discriminators is meaningless. %The diversity of generated sentences is missing

\begin{table*}
%\small 
%\scriptsize
\mysizeTable
\begin{center}
\begin{tabular}{l|l|l}
\Xhline{2\arrayrulewidth}
 \bf Models&\bf Positive $\rightarrow$ Negative &\bf Negative $\rightarrow$ Positive\\ \Xhline{2\arrayrulewidth}
%1
 \bf Source&it has a great atmosphere , with wonderful service .&i have n't received any response to anything .\\ 
 ARAE&it has no taste , with a complete jerk .&i have n't received any problems to please \\
  CAE &it has a horrible atmosphere , with awful service .&i have been pleased with a wonder time .\\
 %CAE$_2$ & it has a horrible atmosphere , and never really .&i have a great waitress and great service .\\ 
 \hline

%2

 \bf Source& the steak was really juicy with my side of salsa to balance the flavor.&my corn beef hash was mushy with uncooked veggies .\\
 ARAE&the steak was really bland with the sauce and mashed potatoes .& my boyfriend brought shrimp and eggs with mushrooms \\
  && were amazing .  \\

  CAE & the steak was really dry with my sauce on the salsa .&my beef hash was juicy and tender .  \\
 %CAE &the steak was really overcooked with my mouth about the flavor&it was amazing !\\ 
 %&to be desired .&
 %\\
 \hline

\Xhline{1.5\arrayrulewidth}
%\multicolumn{3}{c}{}\\
\end{tabular}
\vspace{2mm}
\begin{tabular}{l|l|l}
% \end{tabular}
% % \end{center}
% % \caption{\label{table_new1} Sentiment-transferred examples from the Yelp dataset. } 
% % \end{table*}

% % \begin{table*}
% % %\scriptsize
% % %\small
% % \mysizeTable
% % %\footnotesize
% % \begin{center}
% \begin{tabular}{l|l|l}
\Xhline{2\arrayrulewidth}
 \bf Models &\bf {Entertainment $\&$ Music} $\rightarrow$ {Politics $\&$ Government } &\bf  {Politics $\&$ Government } $\rightarrow$  {Entertainment $\&$ Music}\\ \Xhline{2\arrayrulewidth}

%#1
 %   \bf Source& from big brother , what is the girls name who had $[$unk$]$ in her apt ?&who do you think will be next president of the u . s ? \\
 % ARAE& is big brother in the $[$unk$]$ what do you think of her ?&who do you think will be next to u ? \\
 % CAE& in the middle east , what is the worst $[$unk$]$ in the u . s . military ?&who do you think will be next american idol ? \\ \hline
 % \bf Source & how do you publish a song ?&which is the country with top development indicators ?  \\ 
 % ARAE&how do you react a song ? &which is the best country with led zepplin ?  \\
 % CAE&how do you handle a war ?  &which is the best way to create mp3s ?\\ \hline  

  \bf Source & how do you publish a song ?&who do you think will be next president of the u . s ?  \\ 
 ARAE&how do you react a song ? &who do you think will be next to u ?  \\
 CAE&how do you handle a war ?  &who do you think will be next american idol ?\\ \hline  

    %\bf Source& what is your favorite movie quote ? &who do you think will be next president of the u . s ? \\
 %ARAE& if the majority of america are so $<$unk$>$ on us and all society ?&who do you think will be next to u ? \\
 %CAE& what is your opinion on iraq ?&who do you think will be next american idol ? \\ \hline

 %\bf Source&do you know a website that you can find people who want to join bands ? &can anyone tell me how i could go about interviewing north vietnamese soldiers ?\\
%#2

% %#3
%   \bf Source&where is the tickets for the filming of the suite life of zack and cody ? &republicans : would you vote for a cheney / satan ticket in 2008\\
%  ARAE&where is the best place of the navy and the senate of the union ? &guys : would you rather be a good movie ?\\
%  CAE&where is the nearest place of the royal family of the president of london ? &which celebrity would you vote for a contestant on thanksgiving ?\\ \hline
 
 %#4
 %  \bf Source &do you know a website that you can find people who want to  &can anyone tell me how i could go about interviewing north\\
 %  &join bands ?& vietnamese soldiers ?\\
 % ARAE&do you think that you can find a person who is in prison ? &can anyone tell me how i could find out about my parents ?\\
 % CAE&do you know what a website that you can get to join the military ? &can anyone tell me how i could go to find other people's shoes ?\\ \hline

 \bf Source &do you know a website that you can find people who want to  &what is the gdp of the us currently ? \\
  &join bands ?& \\
 ARAE&do you think that you can find a person who is in prison ? &what is the largest of the us currently ?\\
 CAE&do you know what a website that you can get to join the military ? &what is the name of the band currently ? \\ 
 %\hline
 
 % \bf Source &where can i find a cd by joan jett / $<$unk$>$ love rock n roll ? &how does the united states sanction cuba , but can have a base  \\ 
 %  & & in guantanamo , cuba ? \\ 
% ARAE& how can i find a cd , child support payments for a white n - $<$unk$>$ ?&what is the name of the movie theater , where can i watch the  \\
%  & & movies in atlanta , tx ? \\
% CAE&where can i find a copy of my birth certificate by british army $<$unk$>$ ? & does anyone know where i can find the song , " go on a river " , ?\\ \hline 
 
 %#5
%   \bf Source & where can i find a cd by joan jett / $<$unk$>$ love rock n roll ? &what is the gdp of the us currently ? \\ 
%  ARAE&how can i find a cd , child support payments for a white n - $<$unk$>$ ?&what is the largest of the us currently ?  \\
% % CAE& where can i find a copy of my birth certificate by british army $<$unk$>$ ? & do you have a favorite song and artist ?\\
% CAE & where can i find a copy of my birth certificate from other country  & what is the name of the band currently ? \\
% &$<$unk$>$ canada ?&\\
%  \hline 
 
 %#6
%   \bf Source & what is your favorite movie quote ? &wouldn't it be fun if we the people veto or passed bills ?  \\ 
% ARAE&if the majority of america are so $<$unk$>$ on us and all society ? &it possible or if we $\&$apos;re getting married ? \\
%  CAE& what is your opinion on iraq ? &wouldn't it be fun if we are the people stuck on the floor ?  \\ 
 \Xhline{2\arrayrulewidth}
 
\end{tabular}
\end{center}
\caption{\label{table:sample:yahoo} Style-transferred examples. Upper: from the Yelp dataset. Bottom: from the Yahoo dataset.} % Example of topic transfer on
\end{table*}

\subsection{Analyses}
\subsubsection{Style-transferred sentences}
We display some examples %from the Yelp and Yahoo test sets 
to look into the differences between the CAE and previous approach ARAE
%~\cite{pmlr-v80-zhao18b}   
in Table%~\ref{table_new1} (Yelp) and
~\ref{table:sample:yahoo} 
%(Yahoo)
. In the first example in Table~\ref{table:sample:yahoo}, we can clearly see that CAE correctly detects the sentiment words ``great" and ``wonderful" and successfully transfers the positive sentiment to the negative sentiment by changing the two words into negative words ``horrible" and ``awful". It is worth noting that CAE preserves the substance of the source sentence during the successful style (sentiment) transfer. In contrast,  ARAE fails to keep the background and the major content of original sentences when it struggles to change the style of them not only in the first example but also in other examples.
% in Table~\ref{table_new1}. %proposed

Examples from Yahoo dataset
%in Table~\ref{table:sample:yahoo} 
again demonstrate the advantages of CAE in both style transfer and content preservation over ARAE. It can be obliviously found that CAE is able to learn the patterns of the original questions and to change the topic from ``Entertainment $\&$ Music" to ``Politics $\&$ Government" or vice versa in the frame of the learned patterns. 
\subsubsection{Comparison with the nearest neighbour sequences from training data}
%\todo{why not use the same font size in table 5 and 6} meaningful and useful
We compared the transferred sequences with the nearest-neighbor sequences from training data based on Jaccard distance (word-level intersection over union for two sequences). The results are listed in Table~\ref{table_new2}. The transferred sentences are very different from the retrieved nearest sentences in both syntax and semantics. Additionally, they are also very fluent. This suggests that CAE is capable of learning the style knowledge from training instances and generalizing the learned knowledge to generate style-transferred sentences from given source sentences. 
% For instance, the transferred sentence, ``it has a horrible atmosphere , with awful service .", is different in both the syntax and sentimental words from the retrieved nearest sentence: ``horrible atmosphere , horrible service .". % generating new sequences. for generation

\begin{table*} [h]
%\footnotesize
%\footnotesize
\mysizeTable
\begin{center}
\begin{tabular}{l|l}
\Xhline{2\arrayrulewidth}
\bf Transferred sentences& \bf Nearest neighbour in training data \\ \Xhline{2\arrayrulewidth}
%horrible condition of the place . &the layout of the place is horrible .\\ \hline
 it has a horrible atmosphere , with awful service . 
 & horrible atmosphere , horrible service .\\ \hline 
 definitely a waste of time for sushi in las vegas !&best sushi in las vegas !\\ \hline 
% you could see the best care of the food !&the best of the best !\\ \hline 
%there are a few tables ago and i was eating here on a saturday .&i remember eating here a few years ago and the food was bland .\\ \hline 
%&was bland .\\ \hline 
%he was very personable and the bartender was extremely helpful and knowledgeable .&he was very personable and helpful .\\  \hline
%helpful and knowledgeable .&\\ \hline
%helpful and knowledgeable .&\\ \hline
%you could not see the bobbie at all !&see you all soon ! \\ \hline 
% we came back at the register and it was not very pleased .&came back at 5:30 pm and the wheel was not replaced . \\ \hline 
%their menu is basic , even with thai food .&instead , their food is bland . \\ \hline
%he was very friendly and took us on a weekday !&he was very friendly and did a great job ! \\ \hline 
%no wonder it was closed .& no wonder it was $ \_num\_$ .\\ \hline 
% we came on the recommendation and a side of the worst experience .& we came here on the recommendation of a friend .\\ \hline 
the steak was really dry with my sauce on the salsa .&the philly was dry with no sauce .  \\ \Xhline{2\arrayrulewidth}
\end{tabular}
\end{center}
%\vspace{-2mm}
\caption{\label{table_new2} Comparison between transferred sentences generated by CAE and nearest-neighbour sentences.
% found in the Yelp restaurant reviews sentiment transfer training data. 
} 

\end{table*}

\section{Conclusion}
%\section{CONCLUSION}

We have presented a novel approach, CAE, to unsupervised text style transfer from non-parallel text. We learn latent representations for sequences in different styles with LSTM autoencoders. The learned representations are transferred from their original style to another style via adversarial transfer networks. The transfer networks are equipped with a cycle-consistent constraint to guarantee content preservation during style transfer. Experiments and analyses on the Yelp and Yahoo datasets sufficiently demonstrate the powerful style transfer ability of CAE with good fluency and content preservation against previous methods.

% {\small \begin{table}
% \small
% \begin{center}
% \begin{tabular}{ccccc}
% \Xhline{2\arrayrulewidth} \bf Models & \bf Transfer& \bf BLEU  &\bf  PPL  &\bf RPPL \\ \Xhline{2\arrayrulewidth}
%  AE &$62.0\%$& 32.2&62.7&57.3  \\ \hline 
%  {Template} & $85.3\%$ & 58.9 &137.5&178.4\\ \hline 
%  {Cycled RL}    &  $75.0\%$ & 20.2 & {37.5}& {305.7} \\ \hline 
% {Cross-aligned AE} &$73.4\%$ &11.5&36.5& 79.3\\ \hline
%  ARAE  & $77.7\%$& 21.1& 34.4& 53.7\\ \Xhline{2\arrayrulewidth}
%  %M2Pyfv7 
%  CAE&$ 80.2\% $  &  { 21.2} & 24.9 & 54.9\\
% \Xhline{2\arrayrulewidth}
% \end{tabular}
% \end{center}
% \caption{\label{table_6}Quantitative comparisons on the Yahoo QA topic transfer dataset.%QA topic transfer %{\color{red} The result of cycled RL seems unstable. The intern runs two times, and the result varies a lot for same init param. Two evaluations all get high PPL and RPPL. Not sure about the reason.}%CAE is better than ARAE and Cross-aligned AE based on {\bf Transfer}, {\bf PPL} and {\bf BLEU}.      ({\color{red}just 1 ep})   {\color{red}($27\% [unk]$ in $\tilde{x}$)}
% }
% \end{table}}

% \section*{Acknowledgements}

% The acknowledgements should go immediately before the references.  Do
% not number the acknowledgements section. Do not include this section
% when submitting your paper for review.

\bibliographystyle{coling}
\bibliography{coling2020}

\end{document}